\documentclass[journal]{IEEEtran}

\usepackage{enumitem}
\usepackage{amsfonts, amsmath}
\usepackage[mathscr]{euscript}
\usepackage{graphicx}
\usepackage[noadjust]{cite}
\usepackage{array}
\usepackage[caption=false,font=normalsize,labelfont=sf,textfont=sf]{subfig}
\usepackage{multirow}
\usepackage{afterpage}
\usepackage{fixltx2e}

\begin{document}
\title{Deep Motion Blur Removal Using Noisy/Blurry Image Pairs}

\author{\IEEEauthorblockN{Shuang~Zhang\IEEEauthorrefmark{1}} \thanks{\IEEEauthorrefmark{1} Address all correspondence to: Shuang Zhang, Email: szhang15@nd.edu},~\IEEEmembership{Member,~IEEE,}
        Ada~Zhen,~\IEEEmembership{Member,~IEEE,}
        and~Robert~L.~Stevenson,~\IEEEmembership{Member,~IEEE}
\thanks{Shuang Zhang, Ada Zhen and Robert Stevenson are with the Department
of Electrical Engineering, University of Notre Dame, Notre Dame,
IN, 46556.}
}

\maketitle

\begin{figure*}[!t]
\centering
	\includegraphics[width=1.9\columnwidth]{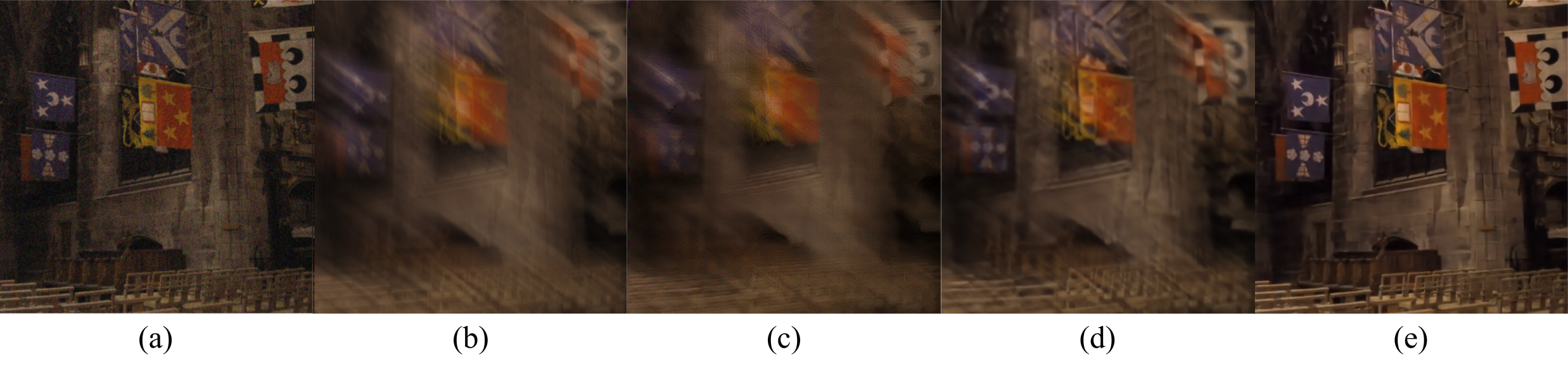}
	\caption{Deblurring results of an image from K{\"o}hler dataset \cite{Kohler:2012} degraded by a large blur kernel. (a) Input noisy images of the proposed method. (b) Input blurry images. (c) DeblurGAN \cite{DeblurGAN:2018}. (d) SRN \cite{SRN:2018}. (e) Proposed DeblurRNN. }
\label{Fig:01}
\end{figure*}

\begin{abstract}
Removing spatially variant motion blur from a blurry image is a challenging problem as blur sources are complicated and difficult to model accurately. Recent progress in deep neural networks suggests that kernel free single image deblurring can be efficiently performed, but questions about deblurring performance persist. Thus, we propose to restore a sharp image by fusing a pair of noisy/blurry images captured in a burst. Two neural network structures, DeblurRNN and DeblurMerger, are presented to exploit the pair of images in a sequential manner or parallel manner. To boost the training, gradient loss, adversarial loss and spectral normalization are leveraged. The training dataset that consists of pairs of noisy/blurry images and the corresponding ground truth sharp image is synthesized based on the benchmark dataset GOPRO. We evaluated the trained networks on a variety of synthetic datasets and real image pairs. The results demonstrate that the proposed approach outperforms the state-of-the-art both qualitatively and quantitatively.
\end{abstract}


\section{Introduction}
\IEEEPARstart{M}{otion} blur is one of the most pronounced artifacts in photos captured by hand-held cameras. Camera shake due to long exposure time requirements under low-light conditions and fast object motions in a dynamic scene both contribute to the blur artifact. One popular strategy to avoid motion blur is to decrease exposure time and increase sensor sensitivity setting (ISO). The captured image appears to be sharp but noisy and less colorful. Thus, burst denoising techniques that merge a burst of such frames \cite{Ringaby:2014,hasinoff2016burst,InertiaUKF:2018} were proposed for noise reduction and fine detail disclosure. Another strategy is to directly solve the ill-posed image deblurring problem which is formulated as disentangling the latent sharp image from the unknown blur kernel. The early work parametrized blur models with simple assumptions on the sources of blur. More recent research explored spatially variant blur kernels that are approximated by local motion vectors or a sequence of homographies. Though various blur models have been proposed, the blur in real images is far more complex than any parameterized model. In addition, the following iterative deconvolution process makes them difficult to be employed on mobile platforms. 

Thanks to the progress in deep learning, Convolutional Neural Networks (CNN) have demonstrated their power for the image deblurring problem. Compared to conventional approaches, its short processing time and generalization to any blur type are very promising. The pioneering work \cite{Sun:2015} and \cite{Schuler:2016} are not fully free from the standard deblurring pipeline. They trained neural networks to predict motion blur kernel in spatial or frequency domain and restored the sharp image by time-consuming deconvolution. Recently, the end-to-end deblurring networks have drawn much attention. These techniques produce the latent sharp image from a blurry one in one pass without explicitly estimating blur kernels, thus intrinsically
avoiding artifacts and distortions caused by erroneous kernels. A multi-scale network \cite{Nah:2017} was built by transferring the traditional coarse-to-fine scheme to CNN. Its performance is significantly improved in SRN \cite{SRN:2018} that embedded Recurrent Neural Networks (RNN) into the multi-scale structure. The development of Conditional Generative Adversarial Nets (CGAN) inspired another type of deep deblurring network called DeblurGAN \cite{DeblurGAN:2018}. To alleviate the pattern artifacts, dark channel prior was incorporated into loss function and the residual nets of the DeblurGAN was replaced with the light-weighted U-net \cite{DarkChannelGAN:2019}. 

However, the poor performance in challenging applications is still an issue in those networks. For example, the multi-scale network proposed by SRN \cite{SRN:2018} fails to handle the large blur kernel as shown in Fig. \ref{Fig:01}. Although SRN adopted three scales of neural networks, it is not enough for such large blur kernel. The design of three scales is a trade-off in terms of network size and capability to remove significant blur effects. Similar performance can also be seen in GAN-based networks.

Observing that the noisy sharp image is a very good initial approximation to the latent clean image, we propose to use the noisy/blurry image pair captured in a burst as the input to the network. The noisy image preserves the large-scale structures while the blurry image has the correct color and a high Signal-to-Noise Ratio (SNR). Our goal is to produce a high quality
image by combining two degraded images. This technique has been considered in conventional deblurring approaches \cite{Yuan:2007}\cite{Zhang:2013}, but it has not been used with neural networks yet. In this paper, we present two neural networks, DeblurRNN and DeblurMerger, to exploit the complementary information in both images. DeblurRNN consists of two encoder-decoder structures in sequence, one for denoising and another for deblurring, which are linked by a hidden state, located between the encoder and the decoder. A discriminator is added to the deblurring net to help train the deblurring net adversarially. DeblurMerger uses parallel branches to extract different types of information from different inputs, learns to merge them in the middle layers and to decode them into a sharp image. Both networks can effectively tackle different sizes of blurs caused by different sources. The performance has been verified by both the synthetic datasets and the real image pairs. 

The main contributions of this paper include the following:
\begin{itemize} 
\item We propose two multi-image deblurring networks with well-designed loss functions to benefit from both the noisy image and the blurry image.
\item We explore different ways of combining the information of the noisy and blurry image pair.
\item We propose an approach to generate realistic training and evaluation data that consists of noisy/blurry image pairs and the corresponding ground-truth based on GOPRO dataset. 
\item We compare the proposed network with different capture strategies including single image deblurring, single image denoising and burst denoising.
\end{itemize}

In Section \ref{sec:related_work}, previous related works using traditional techniques and recent neural networks are discussed. The details of the proposed network are described in Section \ref{sec:approach}, which includes network architectures in Section \ref{sec:architecture}, loss functions in Section \ref{sec:loss_func} and data generation in Section \ref{sec:data}. Section \ref{sec:experiment} compares the proposed approach with both single image deblurring/denoising schemes and multiple image denoising methods on synthetic datasets and real image pairs. Finally a conclusion is drawn in Section \ref{sec:conclusion}.

\section{Related Work}
\label{sec:related_work}

\subsection{Multiple Images Deblurring and Denoising}
Images with different exposure settings or different blur directions are used for robust kernel estimation. Yuan \textit{et al.} \cite{Yuan:2007} adopted noisy/blurry image pairs to estimate translational blur kernel. Zhang \textit{et al.} \cite{Zhang:2013} presented a multi-image restoration method which can achieve joint alignment and deblurring. Cho \textit{et al.} \cite{cho2012registration} transformed the spatially-varying kernel estimation into a registration problem using two blurred images. Ben-Ezra \textit{et al.} \cite{ben2003motion} built a hybrid imaging system where the primary camera captures the blurred image and the second camera records a low-resolution video used for motion estimation. Bar \textit{et al.} \cite{bar2007variational} first considered multiple motion-blurred images to handle object motion blur by developing a unified framework of image segmentation and restoration. Wulff \textit{et al.} \cite{wulff2014modeling} applied the fully generative model on a blurred video and estimated linear blur kernels for both foreground and background. Zhen \textit{et al.} \cite{zhen2018inertial} proposed an alternate-exposure capture strategy and simultaneously recorded inertial sensor readings to jointly solve depth estimation, motion segmentation and deblurring.

Denoising can also benefit from multiple images. The classic multi-image denosing techniques such as VBM4D \cite{maggioni2012video} and non-local means \cite{buades2005non} group similar patches
across time and jointly filter under the assumption
that multiple noisy observations can be averaged to better
estimate the true underlying signal. To make it feasible on mobile platforms, Hasinoff \textit{et al.} \cite{hasinoff2016burst} first aligned image patches in a coarse-to-fine scheme and then performed a pairwise robust fusion. Recently, Mildenhall \textit{et al.} \cite{mildenhall2018burst} proposed a convolutional neural network architecture conceptually similar to non-local means to predict spatially varying kernels that can both align and denoise frames.

\subsection{GAN-based Deblurring Networks}
Generative Adversarial Nets (GAN) \cite{GAN:2014} are deep neural net architectures composed of two networks, a generator $G$ and a discriminator $D$. A fake sample $G(z)$ is generated by the generator from input noise $z$ and discriminator $D$ aims to estimate the probability that the fake sample is from training data rather than generated by generator. These two networks are iteratively trained until the discriminator cannot tell if the sample is real or fake. This process can be summarized as a two-player min-max game. Mirza \textit{et al.} \cite{CGAN:2014} extended GAN into a conditional model, Conditional Generative Adversarial Net (CGAN) in (\ref{eq:CGAN}), which feeds auxiliary information to both generator and discriminator to direct the data generation process.
\begin{align} \label{eq:CGAN} \nonumber
\min_G \max_D &~ \mathbb{E}_{\bar{x}\sim \text{P}_{\text{data}}(\bar{x})} \left[ \log D(\bar{x}, y)\right] \\
&+ \mathbb{E}_{z\sim \text{P}_{z}(z)} \left[ \log (1-D(G(z,y),y)) \right],
\end{align} 
where $\text{P}_{\text{data}}$ and $\text{P}_{z}$ respectively denote distribution over training data $\bar{x}$ and input noise $z$, and $y$ is the auxiliary information. 

DeblurGAN is the first end-to-end deblurring network proposed based on the structure of CGAN and the blurry image is regarded as the auxiliary information. Its generator contains two strided convolution layers, nine residual blocks and two transposed convolution layers. DeblurGAN replaced the adversarial loss shown in (\ref{eq:CGAN}) with WGAN loss \cite{WGAN:2017} as the critical function to stabilize the training. In addition, it adopted the perceptual loss \cite{Perceptual:2016} instead of the classical L1/L2 norm distance in the generator loss function. Although DeblurGAN achieves higher PSNR and SSIM than previous deblurring networks like \cite{Sun:2015}, its result still suffers from artifacts that small patterns regularly and repeatedly appear on the deblurred image. 

\begin{figure*}[!t]
	\includegraphics[width=2\columnwidth]{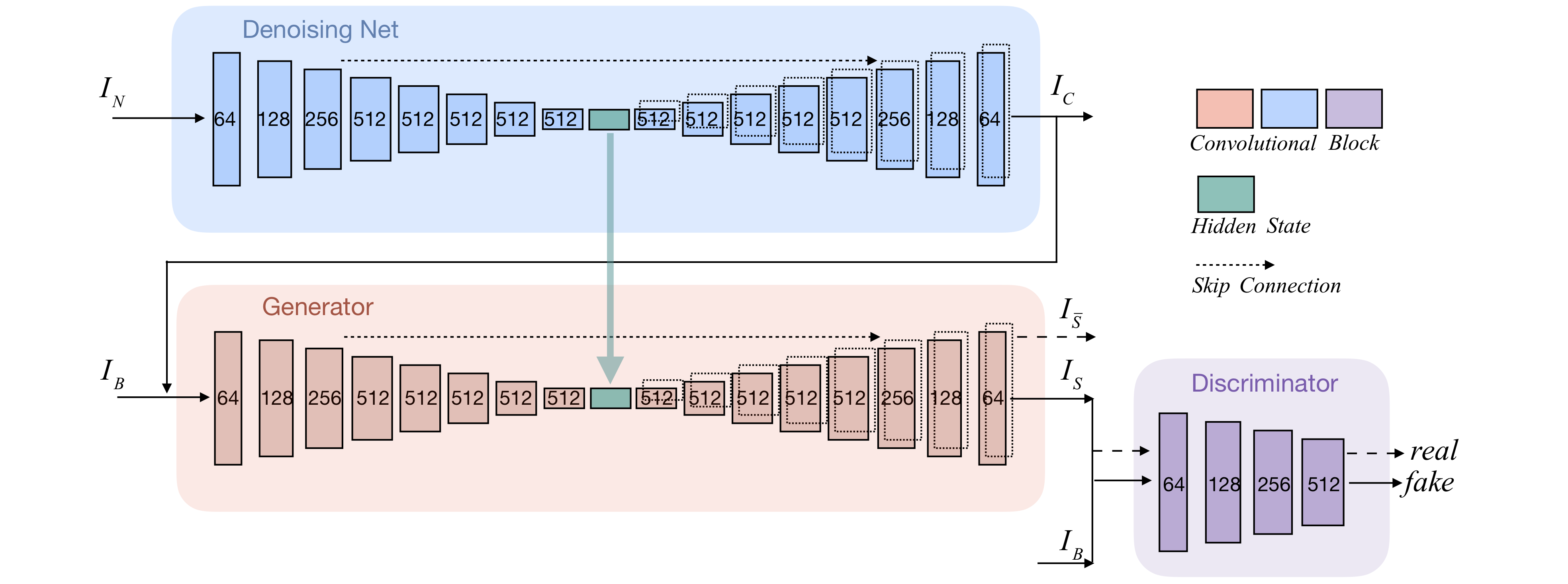}
	\caption{Network structure of DeblurRNN. Except for the first layer of discriminator and generator, each block has a normalization layer. The number inside the rectangle is the size of feature maps. Dotted lines mark the skip connection between the encoder and decoder.}
	\label{Fig:02}
\end{figure*}

An improved network \cite{DarkChannelGAN:2019}, Dark Channel DeblurGAN (DCDeblurGAN), was proposed to suppress the artifacts in DeblurGAN. This network was motivated by the dark channel prior, where the dark channel map of a blurry image is less dark than that of the corresponding latent sharp image \cite{He:2011, Pan:2016}. Pan \textit{et al.} \cite{Pan:2016} has successfully applied the dark channel prior to the image deblurring problem in a conventional optimization framework. Their experiment results suggest the effectiveness of dark channel prior in artifact suppression and detail recovery. The dark channel prior was used in DCDeblurGAN as a part of the generator loss: 
\begin{equation} \label{eq:DC_loss}
\mathscr{L}_{dc} =  \mathbb{E} \left[ ||\mathfrak{D}_c(\bar{x})-\mathfrak{D}_c(G(y,z)) ||_2\right],
\end{equation}
where $\mathfrak{D}_c(p) = \min_{q \in \mathcal{N}(p)} \left( \min_{c \in \{r,g,b\}} I^{c} (q)\right)$ is the dark channel of the pixel location $p$, $\mathcal{N}(p)$ denotes the image patch centered at $p$, and $I^c$ is the $c$-th color channel. DCDeblurGAN also simplified the generator by adopting a U-net structure which consists of an encoder and a decoder \cite{Isola:2017}. The input blurry image goes through a series of convolutional and downsampling layers in the encoder. Once the bottleneck is reached, the transposed convolution layers in the decoder upsample its input feature
maps and finally expand the low resolution image back into a full resolution sharp image.

\subsection{Multi-scale Deblurring Networks}
Multi-scale deblurring was proven to be an effective way to deal with large blur kernels \cite{Wang:2009} in conventional deblurring approaches. A multi-scale deblurring CNN \cite{Nah:2017} was designed in a similar manner. The network starts from the coarse scale of the blurry image. After combining the upsampled deblurred one with the higher resolution blurry image, the concatenated images are fed into the next level of network until the full resolution image is reached. The model is trained with the multi-scale content loss that sums the L2 norm distance between the deblurred ones and sharp images of the corresponding scale.

Nah \textit{et al.} \cite{Nah:2017} also presented a synthetic deblurring dataset (GOPRO) as a benchmark for both training and testing. Different from applying a single blur kernel as the well-known deblurring dataset K{\"o}hler \cite{Kohler:2012}, the GOPRO dataset was generated by integrating a sequence of sharp images in dynamic scenes taken by a high-speed camera. 

Scale-recurrent Network (SRN) \cite{SRN:2018} provides an advanced way of implementing the multi-scale deblurring scheme. In addition to image concatenation, the adjacent scales are also linked by the hidden layer that captures the information of image structures and kernels from the coarser scales. This recurrent structure improves the effectiveness of extracting information from different scales. Instead of using independent parameters for each scale of the cascaded network \cite{Nah:2017}, SRN exploits the same network structure with same weights for each single scale model. Sharing the same set of weights for each scale can not only marginally decrease the number of weights to train but effectively prevent overfitting. 

\section{Proposed Approach}
\label{sec:approach}
In the proposed approach, a pair of images, a noisy image $I_N$ and a blurry image $I_B$,  are captured in rapid succession. The noisy image with high shutter speed and high ISO appears sharp, but it has a very low Signal-to-Noise Ratio (SNR) and the noise is further amplified by the higher camera gain. The blurry image is taken with a slow shutter speed and a low ISO setting. Though it has correct color and high SNR, it is blurry due to camera shakes or object motions. Our goal is to fuse the pair of captured images to reconstruct a high quality image $I_S$ that cannot be obtained by single image denoising or single image deblurring. Two novel networks, DeblurRNN and DeblurMerger, are proposed to combine the information from the noisy and blurry image pair in a sequential or parallel manner. The loss function terms are designed to suppress noise and pattern artifacts. To train the network, a data generation framework that produces a realistic training dataset by simulating different noise profiles and brightness/colors in the image pairs is proposed. Several data augmentation techniques are used to improve the robustness of the trained model.

\subsection{Network Architecture}
\label{sec:architecture}
\subsubsection{DeblurRNN}
In DeblurRNN, the image pair is processed sequentially: denoising first followed by deblurring. Correspondingly, the network is composed of two sequentially cascaded subnets: a denoising net and a deblurring net. 

Fig. \ref{Fig:02} shows the overall architecture of DeblurRNN. The denoising net takes the noisy image $I_N$ as input. Its output $I_C$ is concatenated with the blurry image $I_B$ and fed into the deblurring net. Both nets share the same encoder-decoder structure. The encoder represents the input image with a bottleneck vector and the decoder recovers an image with the same size of input from the bottleneck vector. The encoder consists of a sequence of convolutional blocks and the decoder has a chain of transposed-convolutional blocks. A skip architecture is applied by inserting same size of layers from encoder after each layer of decoder. This skip connection refines the details in the output image by combining deep, coarse, semantic information and shallow, fine, appearance information \cite{Shelhamer:2017}. Dropout is also included in the decoder to avoid over-fitting. As two nets serve different purposes, they are jointly trained using two sets of weights instead of one set of weights shared by two nets. In order to capture the similarity of feature maps of the noisy/blurry pair, ConvLSTM \cite{Shi:2015}, commonly seen in RNN, is inserted between the encoder and the decoder of both nets and the hidden state in the ConvLSTM flows across two nets.

A discriminator is added to form a CGAN along with the deblurring net to boost its training. The discriminator consists of a series of convolutional blocks, and its output is a scalar, followed by a sigmoid function to estimate the probability that the deblurring net output $I_{\bar{S}}$ is the real ground truth image rather than fake sharp image (deblurred image).

Each convolutional (transposed-convolutional) block contains a convolutional (transposed-convolutional) layer with stride $=2$ and kernel size $=5$, a batch normalization layer \cite{Ioffe:2015} and an activation function LeakyReLu \cite{Xu:2015}. A spectral normalization is also inserted before regular batch normalization in each block. The spectral normalization was recently proposed by Miyato \textit{et al.} \cite{Miyato2018} to stabilize the training of the discriminator. The spectral normalization is directly applied on the weights, while batch normalization normalizes output feature maps of the preceding block. Zhang \textit{et al.} \cite{SelfAttention:2018} extended the use case of spectral normalization to the generator. Their results empirically demonstrate that spectral normalization in the generator can prevent the blowing of parameter magnitudes and avoid unusual gradients with low computational cost.

\begin{figure}[ht]
    \centering
	\includegraphics[width=1\columnwidth]{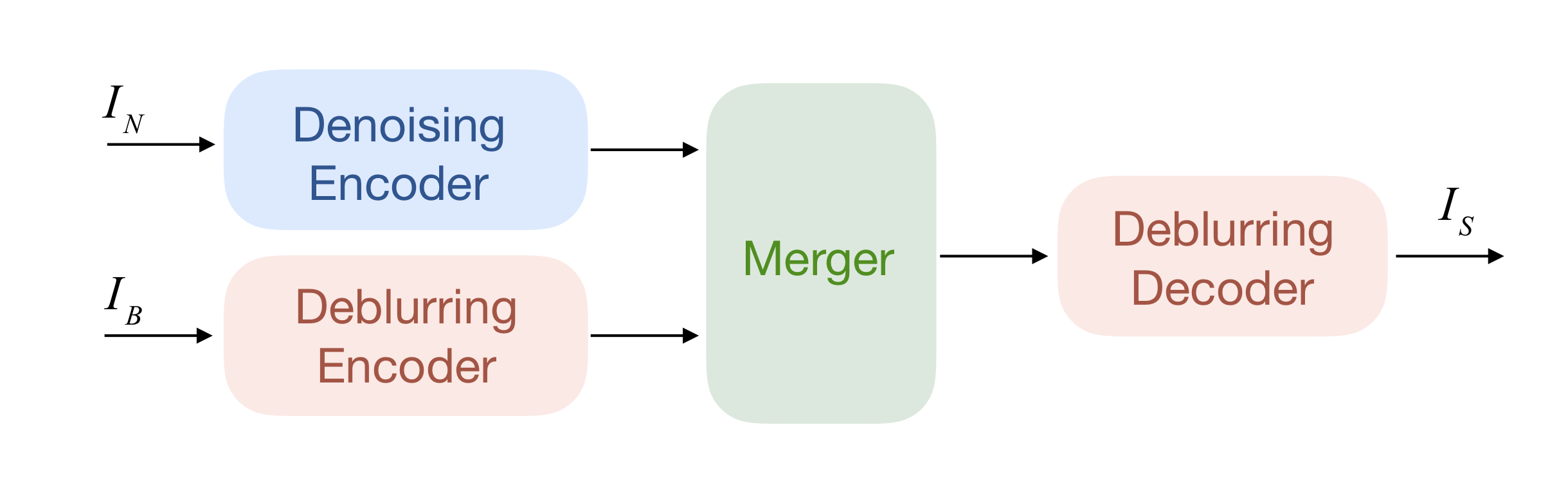}
	\caption{The generator of DeblurMerger.}
	\label{Fig:03}
\end{figure}

\subsubsection{DeblurMerger}
Instead of deferring the information from the blurry image, another network structure called DeblurMerger is proposed to combine the noisy/blurry pair information in parallel. The discriminator described in DeblurRNN is also leveraged in DeblurMerger to boost the training of the entire network.   

The overall architecture of the generator can be conceptually divided into three components (Fig. \ref{Fig:03}):
encoder, merger and decoder.  Since the inputs was differently-exposed images,
separate encoders are used to extract different types of information from
different inputs. After extracting the features, the network learns to merge them
in the middle layers, and to decode them into the final output. The two encoders and one decoder in DeblurMerger are the same as the ones in DeblurRNN. Concatenating the bottlenecks of the two input images in the merger is a simple but effective way to benefit the information of both images \cite{NonlocalNN:2018} as the response at a position is computed as a weighted sum of the features at all positions in the input feature maps.

\subsection{Loss Function}
\label{sec:loss_func}
The loss function of the deblurring generator in DeblurRNN and DeblurMerger is defined as a combination of adversarial loss, content loss, gradient loss and image prior loss:
\begin{equation}
\mathscr{L}_{deblur} = \mathscr{L}_{adv}+\lambda_{c}\mathscr{L}_{c}+\lambda_{grad}\mathscr{L}_{grad}+\lambda_{dc}\mathscr{L}_{dc}.
\end{equation}

The adversarial loss function is essential for the training of both deblurring generator and discriminator:
\begin{equation} \label{eq:d_loss}
\mathscr{L}_{adv} = \mathbb{E}_{\bar{x}, y} \left[ \log D(\bar{x}, y)\right] + 
\mathbb{E}_{y,z} \left[ \log (1-D(G(z,y),y)) \right],
\end{equation} 
where $y$ and $\bar{x}$ denote the blurry and ground-truth sharp image, respectively, $z$ is the random noise. Equation (\ref{eq:d_loss}) is adopted for the training of discriminator $D$, while only the second term is involved in the training of generator $G$. 

The content loss $\mathscr{L}_{c}$ is the difference between the generator output $\hat{x}=G(z,y)$ and ground-truth sharp image $\bar{x}$ in $L1$ norm:
\begin{equation} \label{eq:L1_loss}
\mathscr{L}_{c} =  \mathbb{E} \left[ ||\bar{x}-\hat{x} ||_1\right].
\end{equation}
Though the $L1$ loss or even more delicate loss (such as perceptual loss) is proven to be effective to produce sharp results, using it as the optimization target leads to pattern artifacts on generated images. The gradient loss and the image prior loss aim to suppress the pattern artifacts. The gradient loss $\mathscr{L}_{grad}$ \cite{Jiao:2017} is
\begin{align} \label{eq:L1_grad} 
\mathscr{L}_{grad} = &~ \mathbb{E} \left[ \left||\nabla_h (\bar{x})- \nabla_h (\hat{x}) \right||_1+\left||\nabla_v (\bar{x})- \nabla_v (\hat{x}) \right||_1 \right],
\end{align}
where $\nabla_h$ and $\nabla_v$ denote the horizontal and vertical gradient operators which are approximated by applying the Sobel filter. The dark channel loss $\mathscr{L}_{dc}$ defined in (\ref{eq:DC_loss}) is employed as image prior loss.  

The denoising net of DeblurRNN is trained using only the content loss
\begin{equation}
\mathscr{L}_{denoise} = \mathbb{E}\left[ ||\bar{x}-\tilde{x} ||_1\right],
\end{equation}
in which $\tilde{x}$ represents the output of the denoising net. As the deblurring net produces the final output, less constraints are imposed on the denoising net.

\begin{table}[!h]
\caption{Loss Functions for Each Net.}
\label{tab:loss_function}
\centering
\begin{tabular}{c||c|c|c|c}
\noalign{\hrule height 1pt}
Nets &  $\mathscr{L}_{adv}$ & $\mathscr{L}_{c}$ & $\mathscr{L}_{grad}$ & $\mathscr{L}_{dc}$\\
\noalign{\hrule height 1pt}
Denoising & ~ & $+\lambda_{c1}$ & ~ & ~\\
\hline
D of DeblurRNN & $+1$ & ~ & ~ & ~\\
\hline
G of DeblurRNN & $+1$ & $+\lambda_{c2}$ & $+\lambda_{grad}$ &  $+\lambda_{dc}$\\
\hline
D of DeblurMerger & $+1$ & ~ & ~ & ~\\
\hline
G of DeblurMerger & $+1$ & $+\lambda_{c2}$ & $+\lambda_{grad}$ &  $+\lambda_{dc}$\\
\noalign{\hrule height 1pt}
\end{tabular}
\end{table}

For DeblurRNN, the training is performed in the order of the denoising net $\mathscr{L}_{denoise}$, the discriminator $\mathscr{L}_{adv}$ and the deblurring generator $\mathscr{L}_{deblur}$. As for DeblurMerger, the training order is the discriminator $\mathscr{L}_{adv}$ followed by the generator $\mathscr{L}_{deblur}$. Table. \ref{tab:loss_function} summarizes the usage of loss functions mentioned above, where $\lambda_{c1} = \lambda_{c2} = \lambda_{grad} = 50$ and $\lambda_{dc} = 250$ in the experiments. "$+\lambda$" means the specific loss function is added after being multiplied with the corresponding coefficient $\lambda$. These parameters are picked following the rule that the value of each loss are in the similar scale in the beginning of the training process.

\subsection{Data Preparation}
\label{sec:data}

In order to train the network, pairs of noisy
and blurry images together with the corresponding sharp
images are required. A data generation framework is proposed to synthesize a realistic training and testing dataset, \textit{GOPRO2}, based on GOPRO\_Large\_all dataset which contains successive sharp frames taken by a GOPRO4 Hero Black camera in more than 30 different scenes. The popular benchmark dataset GOPRO \cite{Nah:2017} was formed by accumulating varying number of successive latent frames in GOPRO\_Large\_all to simulate complex
camera shakes and object motions in real photographs. Same as in \cite{Nah:2017}, GOPRO2 dataset is composed of 3214 sets of noisy/blurry/ground truth images at 1280x720 resolution, in which 2,058 sets are used for training
and the remaining 1,111 sets for evaluation.

Given a sequence of consecutively captured sharp frames, the first image is picked as the ground truth latent image and used for generating the noisy image, and the third to the last frames are averaged to create the blurry image. The second frame is abandoned to mimic the time lag between the noisy image and the blurry image captured by mobile phone cameras. 

The noisy image is generated by scaling intensity of the ground truth image with a random scaling factor $f_{scale}$ which is uniformly sampled from $[0.3, 0.8]$ to account for the exposure difference. The
random exposure ratio encourages the network to learn the color and brightness from the blurry image. In addition, under different exposure ratios, the image is corrupted by noise which is dominated by shot noise or readout noise. In low light conditions, i.e., the scaling factor $f_{scale}$ is less than a threshold, shot noise is the main source
of noise, modeled by a signal-dependent Poisson distribution, while readout noise is dominant in other situations and it is approximated by a signal-independent Gaussian distribution.

The signal-dependent noise is simulated using the model proposed in \cite{Zhang:2009MultipleVI},
\begin{equation} \label{eq:noise_model}
I_{N}(p) = \frac{1}{\sigma_s}{\rm Poisson}(\sigma_s I_{S}(p)),
\end{equation}
where $I_{N}(p)$ denotes the noisy measurement of the true intensity $I_{S}(p)$ at pixel $p$ and $\sigma_s$ is number of unique intensities in the sharp image $I_{S}$. The standard deviation of Gaussian readout noise $\sigma_r$ is uniformly sampled from $[0.05, 0.1]$ to make the proposed network robust
against different strengths of noise. Only readout noise is considered in synthesizing real blurry images and its variance is $\sigma_r^2 / N$ where $N$ is the number of sharp frames used for generating the blurry image.

\section{Experiments}
\label{sec:experiment}
\subsection{Training Details}
The proposed approach is implemented using Tensorflow \cite{TensorFlow:2016}. All the training and testing are performed on a NVIDIA GeForce GTX 1080 Ti GPU. The input training pair is randomly cropped into $256 \times 256$ patches. The noisy images are pre-scaled by an estimated ratio to compensate for the exposure difference between the blurry and noisy images. All trainable variables are initialized using a Gaussian distribution with zero mean and standard deviation $0.02$. During the training process, the Adam solver \cite{Kingma:2015AdamAM} with $\beta_1 = 0.5$ and initial learning rate $lr = 0.0002$ is adopted. For each iteration of optimization, one step of backpropagation is performed on the denoising net (only for DeblurRNN), followed by one step on the deblurring discriminator $D$ and one step on the deblurring generator $G$. The proposed network was trained for 10 epochs within 4 hours.

\subsection{Synthetic Datasets}
The proposed network is compared against recent end-to-end image deblurring networks, DeblurGAN \cite{DeblurGAN:2018} and SRN \cite{SRN:2018}, on synthetic datasets (GOPRO, GOPRO2 and {K{\"o}hler}). Different from GOPRO dataset, {K{\"o}hler} dataset is originally created for single image blind deblurring algorithms. Its ground truth blur kernels are generated by recording and analyzing real camera motions on a robot platform. In the GOPRO and K{\"o}hler datasets, one extra noisy image is synthesized with the method described in Section \ref{sec:data} in order to test them with the proposed network. Both DeblurGAN and SRN take a single blurry image as input. Their results are generated using their official trained model available online\footnote{https://github.com/KupynOrest/DeblurGAN} \footnote{https://github.com/jiangsutx/SRN-Deblur}. 

Fig. \ref{Fig:05} shows deblurred results on GOPRO dataset. The most apparent weakness of DeblurGAN \cite{DeblurGAN:2018} is pattern or checkerboard artifacts which are generated over the entire image. SRN \cite{SRN:2018} can handle different types of blur kernels in general. However, it is still incapable to address blurry images with multiple moving objects or degraded by large blur kernels. On the first row of Fig. \ref{Fig:05}, the little girl and the little boy shown in local images move faster than the other objects in the scene, and SRN result is blurry and lack of details on their faces.

\afterpage{%
\begin{figure*}[!t]
	\includegraphics[width=2\columnwidth]{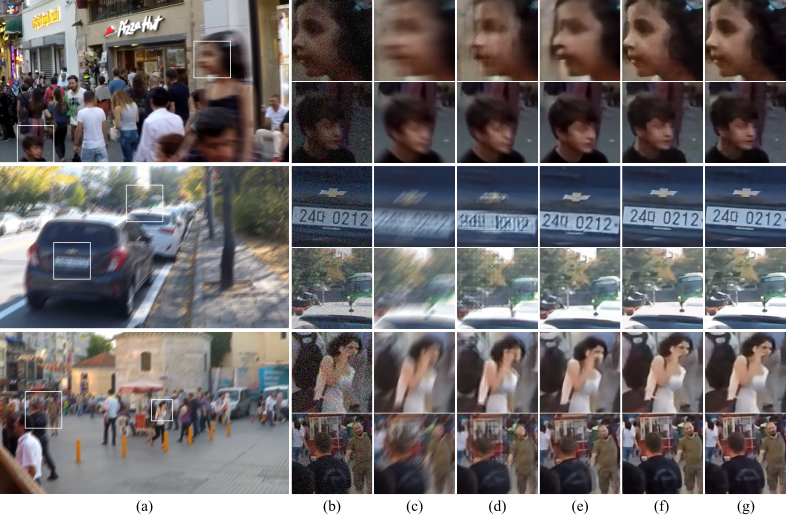}
	\caption{Deblurring results of test images taken from GOPRO dataset. (a) Blurry images with full resolution. (b) Input noisy images of the proposed method. (c) Input blurry images. (d) Results of DeblurGAN \cite{DeblurGAN:2018}. (e) Results of SRN \cite{SRN:2018}. (f) Results of proposed DeblurMerger. (g) Results of proposed DeblurRNN. }
	\label{Fig:05}
\end{figure*}

\begin{figure*}[!b]
	\includegraphics[width=2\columnwidth]{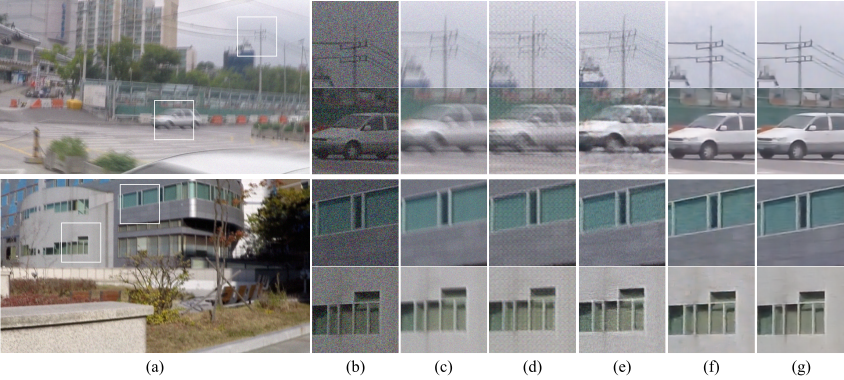}
	\caption{Deblurring results of test images taken from GOPRO2 dataset. (a) Blurry images degraded by noise in full resolution. (b) Input noisy images of the proposed method. (c) Input blurry images. (d) Results of DeblurGAN \cite{DeblurGAN:2018}. (e) Results of SRN \cite{SRN:2018}. (f) Results of proposed DeblurMerger. (g) Results of proposed DeblurRNN. }
	\label{Fig:06}
\end{figure*}
\clearpage
}

Compared to GOPRO dataset, the proposed GOPRO2 dataset adds noise on the blurry image, which is even more challenging for DeblurGAN and SRN. As shown in Fig. \ref{Fig:06}, both the two networks perform much worse than on GOPRO dataset. They fail to remove most of the blur and show different kinds of artifacts and higher noise. The proposed DeblurRNN and DeblurMerger demonstrate more robustness to the noise on these blurry images. Although subtle to notice the result of DeblurRNN is slighter better than the result of DeblurMerger. For example, the window frame is better defined in DeblurRNN's result on the bottom row of Fig. \ref{Fig:06}.

\begin{table}[!h]
\caption{Average PSNR and SSIM for GOPRO, GOPRO2 and K{\"o}hler dataset. }
\centering
\label{tab:sythetic_deblur}
\begin{tabular}{c||c|c|c|c|c}
\noalign{\hrule height 1pt}
\multirow{2}*{Dataset}  & \multirow{2}*{Metrics}  & Kupyn   & Tao  & \multirow{2}*{Merger}& \multirow{2}*{RNN}\\
~ &  ~ & \cite{DeblurGAN:2018} & \cite{SRN:2018} & ~\\
\noalign{\hrule height 1pt}
\multirow{2}*{GOPRO} & PSNR & 26.63 & 30.25 & 31.38 & \textbf{31.98}\\
~     & SSIM & 0.8798 & 0.9422 & 0.9471 & \textbf{0.9533}\\
\hline
\multirow{2}*{GOPRO2} & PSNR & 20.51  & 25.48 & 31.86 & \textbf{32.53}\\
~     & SSIM & 0.7489 & 0.7976 & 0.9531 & \textbf{0.9561}\\
\hline
\multirow{2}*{K{\"o}hler} & PSNR & 25.73 & 26.80 & 31.77 & \textbf{32.90}\\
~          & SSIM & 0.8212 & 0.8579 & 0.9475 & \textbf{0.9592} \\
\hline
\multicolumn{2}{c|}{Time} & 1.8s & 1.6s & \textbf{0.7}s & 1.1s\\
\noalign{\hrule height 1pt}
\end{tabular}
\end{table}

A quantitative comparison of these methods is presented in Table. \ref{tab:sythetic_deblur}. Due to the designed framework, a fair margin on PSNR and SSIM metrics is achieved. In addition, its robustness against noise is also validated by the quantitative numbers. Except for performances, DeblurRNN and DeblurMerger require less processing time for each image.

\subsection{Real Images}
Comparisons are also conducted on real pairs of noisy and blurry images that are captured by a hand-held smartphone camera, LG Nexus 6. In burst mode, the time interval between two shots can be very small. The camera setting of the blurry image is automatic exposure and ISO. The exposure time ratio between noisy and blurry images varies from $1/6$ to $1/10$ and the ISO setting of the noisy image is correspondingly increased to compensate exposure difference. The first row in Fig. \ref{Fig:07} shows one example where the blurry image is corrupted by a large camera motion. Neither DeblurGAN or SRN can restore the sharp image. This is due to the limited number of scales that SRN adopted for consideration of the network size. In the traditional multi-scale deblurring approaches, the coarse-to-fine scheme works based on the assumption that the blurry image at the coarsest level should be a good approximate to the sharp image at the same level. As SRN only uses three scales, this assumption can be violated sometimes. The example on the second row captures a dynamic scene. SRN removes the most motion blur of the car but brings ringing artifacts and some blurs still remain on rear lights. In both cases, however, the proposed network can do a good job at dealing with the complex blurred images, as well as extracting the correct colors from the blurry image.

\begin{figure}[htb]
\centering
	\includegraphics[width=1\columnwidth]{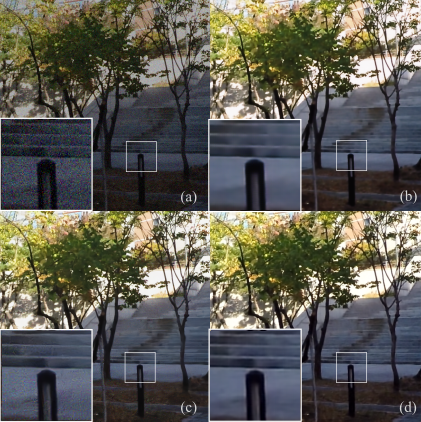}
	\caption{Denoised results of a synthetic noisy image taken from GOPRO2 dataset. (a) Input noisy image. (b) Results of proposed DeblurMerger. (c) Results of Noise2Noise \cite{Noise2Noise}. (d) Results of proposed DeblurRNN. }
	\label{Fig:08}
\end{figure}

\subsection{Comparison with Single Image Denoising Method}
In previous section, the proposed method with the input noisy/blurry image pair achieves better results than single image deblurring method. In this section, the proposed network is compared with the state-of-the-art single image denoising network, Noise2Noise \cite{Noise2Noise}, to verify the necessity of the blurry image for the proposed scheme. The noisy image as input to Noise2Noise is the same noisy image of the noisy/blurry image pair.

\afterpage{%
\begin{figure*}[!t]
	\includegraphics[width=2\columnwidth]{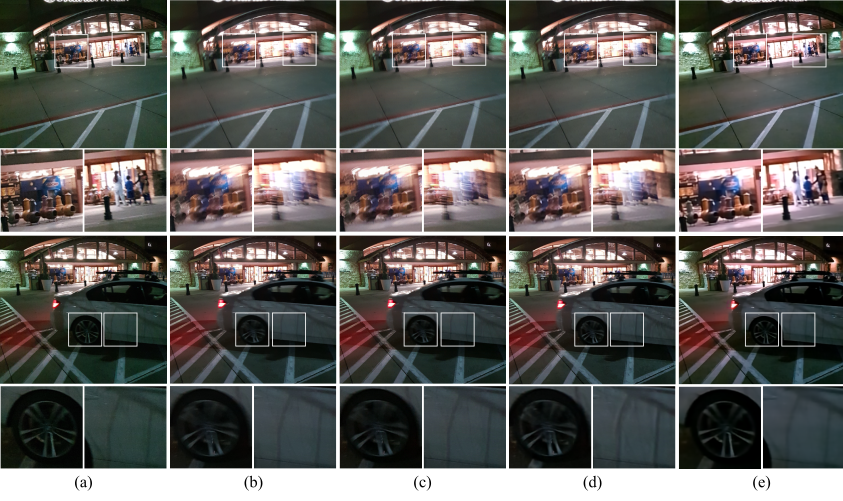}
	\caption{Deblurring results of real images. (a) Input noisy images. (b) Input blurry images. (c) Results of DeblurGAN \cite{DeblurGAN:2018}. (d) Results of SRN \cite{SRN:2018}. (e) Results of proposed DeblurRNN. }
	\label{Fig:07}
\end{figure*}

\begin{figure*}[!b]
	\includegraphics[width=2\columnwidth]{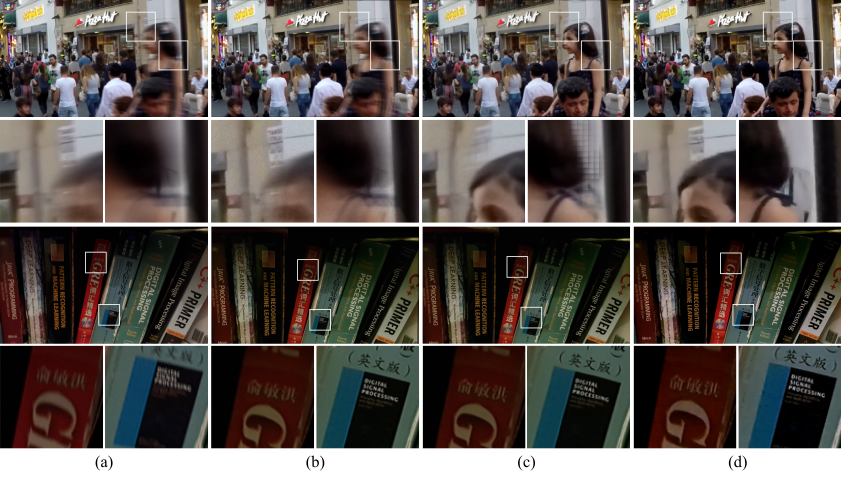}
	\caption{Comparison of burst denoising methods and the proposed method on GOPRO2 example (top) and a real image (bottom). (a) Input blurry images. (b) Results of Ringaby \textit{et al.} \cite{Ringaby:2014}. (c) Results of Hasinoff \textit{et al.} \cite{hasinoff2016burst}.  (d) Results of proposed DeblurRNN. }
	\label{Fig:09}
\end{figure*}

\clearpage
}

Table \ref{tab:sythetic_denoise} presents the quantitative evaluation results of Noise2Noise and the proposed in terms of PSNR, SSIM on the GOPRO2 dataset. The visual comparison is shown in Fig. \ref{Fig:08}. Noise2Noise clearly decreases the noise level. Nevertheless, their result appears more noisy in some local areas compared to the proposed network, which is obtained from the noisy/blurry pair. 

\begin{table}[h]
\caption{Average PSNR and SSIM for GOPRO2 dataset. }
\label{tab:sythetic_denoise}
\centering
\begin{tabular}{c||c|c|c|c}
\noalign{\hrule height 1pt}
\multirow{2}*{Dataset} & \multirow{2}*{Metrics} & Noise2Noise & Deblur & Deblur \\
~ &  ~ & \cite{Noise2Noise} & Merger & RNN\\
\noalign{\hrule height 1pt}
\multirow{2}*{GOPRO2} & PSNR & 30.91 &  31.86 & \textbf{32.53}\\
~     & SSIM & 0.9252 & 0.9531 & \textbf{0.9561}\\
\noalign{\hrule height 1pt}
\end{tabular}
\end{table}

\subsection{Comparison with Burst Denoising Methods}
Burst image denoising is an alternative strategy to the noisy/blurry image pair deblurring. It performs temporal denoising by averaging pixel values of all images, but it is prone to produce significant ghosts in the presence of misalignment or object motions. In this section, two representative burst denoising methods, presented by Hasinoff \textit{et al.} \cite{hasinoff2016burst} and Ringaby \textit{et al.} \cite{Ringaby:2014}, are compared with the proposed method.

Ringaby \textit{et al.} \cite{Ringaby:2014} employs image feature tracking and built-in gyro sensor in smartphone to estimate the relative motion between frames and averages aligned images to generate a clear sharp image. The work of Hasinoff \textit{et al.} \cite{hasinoff2016burst}, also called HDR+, provides a burst denoising pipeline which adopts hierarchy alignment to locally align input noisy frames and applies a tile-based Wiener filter to merge aligned frames in frequency domain. HDR+ avoids introducing ghost effects with more accurate alignment and frequency-domain shrinkage merging. As the official code is not public, the results are produced by an implementation of the previous paper \cite{InertiaUKF:2018}.

To make a fair comparison between the burst denoising and the proposed approach, experiments are conducted under the condition of the same amount of total capturing time. In other words, if the blurry image in the noisy/blurry image pair is generated by 10 latent sharp frames, the input noisy frames to burst denoising is 11 frames (10 for blurry image and 1 for noisy image). Fig. \ref{Fig:09} presents two examples to visually compare the performance. The second row in Fig. \ref{Fig:09} is a real example where only camera misalignment exists. Both of the two burst denoising approaches achieve good denoising level, but the colors of
their results are partially lost. The example on the first row comes from GOPRO synthetic dataset. Not surprisingly, the above two burst denoising methods fail to deal with moving objects in the scene. Though HDR+ performs local alignment to force the registration of the moving object, the motion vectors around the neighborhood of large motion area are not consistent with the motion boundary due to lack of the smoothness constraint in their method. Thus, the image restoring performance on the boundary of moving objects is unsatisfactory. In comparison, the proposed method is able to recover details out of significant noise and does not produce artifacts in the presence of large camera/scene motion.

\section{Conclusion}
\label{sec:conclusion}
In this paper, a framework is proposed to recover a sharp and clean image given a pair of noisy/blurry images. Two network structures, DeblurRNN and DeblurMerger, are presented to extract the complementary information of both input images in a sequential manner and a parallel manner, respectively. A discriminator is appended to the deblurring net to perform adversarial training. A realistic synthetic dataset of blurry/noisy/ground-truth for network training purpose was generated. The experiment results demonstrate that the performance of the trained network exceeds state-of-the-art deblurring networks and single/multiple image denoising methods on both synthetic and real scenes.

\bibliographystyle{IEEEtran}
\bibliography{paper}
\end{document}